\documentclass[a4paper,conference]{IEEEtran}
\IEEEoverridecommandlockouts
\usepackage{verbatim}
\usepackage{amsmath,amssymb,amsfonts}
\usepackage{algorithmic}
\usepackage{url}
\usepackage{graphicx}

\usepackage{textcomp}
\usepackage{xcolor}
\usepackage[hidelinks]{hyperref}
\usepackage{placeins}
\usepackage{dblfloatfix}
\usepackage{enumitem}
\usepackage{float}

\usepackage{array}
\usepackage{ltablex} 
\usepackage{booktabs} 
\usepackage{adjustbox} 
\usepackage{multirow}
\usepackage{graphicx}

\newcommand{\gap}{\hspace{4mm}}

\usepackage{breakurl} 

\usepackage[font=footnotesize]{caption}
  
\begin{document}

\title{Decentralized Weather Forecasting via\\ Distributed Machine Learning and\\ Blockchain-Based Model Validation}


%


\author{%
\IEEEauthorblockN{%
Rilwan Umar,\gap Aydin Abadi,\gap Basil Aldali,\gap Benito Vincent,\gap Elliot A.~J.~Hurley\\
Hotoon Aljazaeri,\gap Jamie Hedley\--Cook,\gap Jamie\--Lee Bell,\gap Lambert Uwuigbusun\\
Mujeeb Ahmed,\gap Shishir Nagaraja,\gap Suleiman Sabo,\gap Weaam Alrbeiqi}
\\
\IEEEauthorblockA{\text{School of Computing, Newcastle University, United Kingdom}}
\thanks{\hspace{-2.7mm}A preliminary version of this paper appears in FLTA 2025.}
\thanks{\hspace{-2.7mm}rilweezy120@gmail.com; aydin.abadi@ncl.ac.uk; B.J.A.Aldali2@newcastle.ac.uk;
B.Vincent3@newcastle.ac.uk; E.A.J.Hurley2@newcastle.ac.uk; H.H.M.Aljazaeri2@newcastle.ac.uk;
J.Hedley\--Cook2@newcastle.ac.uk; J.L.Bell1@newcastle.ac.uk; L.Uwuigbusun2@newcastle.ac.uk;
mujeeb.ahmed@newcastle.ac.uk; shishir.nagaraja@newcastle.ac.uk; S.M.Sabo2@newcastle.ac.uk;
W.Alrbeiqi2@newcastle.ac.uk.}%
}

\maketitle
\thispagestyle{plain}

\begin{abstract}

Weather forecasting plays a vital role in disaster preparedness, agriculture, and resource management, yet current centralized forecasting systems are increasingly strained by security vulnerabilities, limited scalability, and susceptibility to single points of failure. 
%
%
To address these challenges, we propose a decentralized weather forecasting framework that integrates Federated Learning (FL) with blockchain technology. 
 FL enables collaborative model training without exposing sensitive local data, this approach enhances privacy and reduces data transfer overhead. Meanwhile, the Ethereum blockchain ensures transparent and dependable verification of model updates. 
%
%
To further enhance the system's security, we introduce a reputation-based voting mechanism that assesses the trustworthiness of submitted models while utilizing the Interplanetary File System (IPFS) for efficient off-chain storage. Experimental results demonstrate that our approach not only improves forecasting accuracy but also enhances system resilience and scalability, making it a viable candidate for deployment in real-world, security-critical environments.

\end{abstract}




\section{Introduction}
\label{sec: intro}
Weather forecasting is essential for agricultural productivity, disaster preparedness, and economic stability. However, traditional forecasting methods tend to rely on centralized systems. This centralization poses significant risks, including vulnerabilities to data manipulation, privacy breaches, and single points of failure~\cite{bonawitz2019towards}. Centralized Machine Learning (ML) models, despite their high accuracy, are vulnerable to adversarial threats, such as data poisoning, where attackers introduce incorrect data to compromise forecast reliability~\cite{papernot2018sok}. Reliable weather forecasting systems are foundational to sectors like the insurance industry, where the integrity of environmental data directly influences risk assessment and claim processing. In large-scale disaster scenarios, verifiable forecasts and observations can reduce delays, eliminate the need for manual inspections, and streamline high-volume claim settlements. In consequence, adopting decentralized methods for secure and reliable weather forecasting becomes imperative. 

This study adopts a distributed machine learning paradigm that draws upon the core ethos of Federated Learning (FL). Rather than synchronizing training across devices, clients in this system independently retrain models using localized weather data and submit updates to a decentralized validation layer implemented via blockchain. This approach reduces reliance on central servers, minimizes exposure of sensitive data, and aligns with FL's privacy-preserving intent. However, it provides greater architectural flexibility by enabling advanced ML frameworks like TensorFlow and SKtime to operate off-chain (i.e., outside of the blockchain network), thus avoiding the scalability constraints associated with blockchain-native model execution. The integration with decentralized storage via Inter-planetary File System (IPFS) allows for secure, tamper-evident off-chain model storage, overcoming the inefficiencies of storing large ML artifacts on-chain \cite{chen2017under}.

While this distributed approach mirrors FL's regulatory advantages, such as compliance with the General Data Protection Regulation (GDPR) and the European Union (EU) Data Act by preserving local data privacy \cite{bonawitz2019towards}\cite{papernot2018sok}, it improves performance by decoupling model validation from synchronous federation cycles. Nonetheless, it remains susceptible to malicious model submissions and adversarial retraining \cite{sun2021data}, which are mitigated through a smart contract-based consensus mechanism and reputation system. This hybrid architecture, therefore, combines the strengths of decentralized learning and blockchain, optimizing for both resilience and regulatory alignment.
 

Blockchain's core properties, such as distributed consensus and transparent validation, ensure robust data integrity and prevent unauthorized model modifications \cite{bonneau2015sok}. Smart contracts automate verification and validation, ensuring the legitimacy of model updates and user contributions \cite{kim2019blockchained}.
 Although blockchain ensures data integrity and auditability, it does not prevent all attack vectors, such as model poisoning, since adversarial clients can still submit manipulated updates. Therefore, this work implements a consensus-based scoring mechanism to detect and reject such poisoned models through collective validation.

This paper proposes a decentralized weather forecasting framework that integrates ML with blockchain technologies. We introduce a secure, privacy-preserving weather forecasting model, integrated with Long Short-Term Memory (LSTM) networks, enabling collaborative training across distributed datasets. Additionally, we enhance the security of our decentralized FL architecture by implementing blockchain-based smart contracts, providing a decentralized model verification and reputation management mechanism. To address storage demands effectively, we utilize IPFS as a decentralized off-chain storage solution to transcend blockchain scalability and storage constraints. To demonstrate the practical applicability of our proposed system, we also develop a user-friendly backend API, allowing seamless user interaction with the decentralized forecasting platform.

Ultimately,  data-driven approaches foster transparency and trust, as pricing and coverage decisions are based on objective and scientific evidence. However, that is only possible if system operations are verifiable by all parties. This paper, therefore, focuses on security-as-transparency.

The novelty of our work lies in delicately combining federated learning with blockchain-based reputation voting and IPFS-based decentralized storage to enable secure, privacy-preserving, and resilient weather forecasting without relying on a central aggregator.

The remainder of the paper is organized as follows: Section II reviews related work. Section III outlines the problem formulation and key challenges addressed by our approach. Section IV presents the threat model. Section V details the system architecture. Section VI discusses implementation. Section VII evaluates the system’s performance regarding security, resilience, and model accuracy. Section VIII concludes and highlights future work.

\subsection{Key Contributions of the Paper}

This paper makes the following key contributions:
\begin{itemize}
    \item We design a decentralized weather forecasting framework that integrates federated learning with blockchain technology to address core challenges of trust, resilience, and data privacy. By eliminating the need for a central aggregator, our approach enables clients to independently train models on local weather data and submit them to a blockchain-based validation layer. The system ensures architectural flexibility and scalability by using advanced machine learning algorithms and IPFS for decentralized model off-chain storage.
    
    \item We implement and publicly release an open-source prototype~\cite{sourcecode} of the proposed system. The implementation features a reputation-weighted voting mechanism developed in an Ethereum smart contract, which enables secure and transparent model validation. To mitigate Sybil attacks, the system enforces staking-based reputation thresholds and role-based access control. Privacy is further preserved by preventing raw data sharing and securing model artifacts using tamper-evident IPFS identifiers.
\end{itemize}

\section{Background and Related Work}

%
The integration of ML, blockchain, and the Interplanetary File System (IPFS) for decentralized weather forecasting presents a novel approach compared to existing literature. 

The proposed decentralized ML architecture diverges from traditional FL frameworks, such as those proposed by \cite{mcmahan2017communication}, which rely on synchronous aggregation via central servers. Instead, it adopts a fully decentralized paradigm where clients independently retrain models and submit updates to a blockchain-mediated validation layer. This approach aligns with emerging decentralized ML frameworks, such as the exploration of peer-to-peer learning in~\cite{LI2020106854}, but innovates by integrating blockchain for consensus-driven model scoring, a departure from purely gradient-sharing architectures. While \cite{bonneau2015sok}’s foundational work on blockchain security emphasizes immutability and transparency, this study operationalizes these principles to validate ML models, addressing vulnerabilities like adversarial updates through reputation-weighted voting, a mechanism less explored in decentralized ML literature. The use of IPFS for off-chain model storage mirrors decentralized data-sharing frameworks, such as those presented by \cite{batchu2021}. However, its application to weather forecasting with LSTM-based time-series models demonstrates domain-specific novelty. The system’s reliance on untuned LSTM architectures contrasts with optimized forecasting frameworks in \cite{makridakis2018statistical}, raising concerns about predictive accuracy compared to state-of-the-art models. Security mechanisms, such as Sybil-resistant reputation systems, built on trust frameworks of \cite{kang2019incentive}  but inherit unresolved risks highlighted in \cite{fung2018mitigating}, where colluding adversaries might circumvent reputation thresholds. Reliance on smart contracts for governance introduces risks documented in~\cite{LIU2023111775}, such as re-entrancy and front-running, which are partially mitigated but not fully eradicated. 

Some schemes combine blockchain with FL to improve security and transparency in distributed learning systems \cite{wang2023fedchain, ahamed2025flblock}. However, these approaches either focus on generic computation or supply chain use cases.  They do not address forecasting-specific needs like off-chain storage and temporal model verification. Our work extends this by combining IPFS-based storage, LSTM models for time-series forecasting, and smart contract–based model scoring to deliver a secure and scalable weather prediction framework.

\section{Threat Model and Security Assumptions}

\subsection{Threat Model}
We considered a strong threat model that involved both insider threats and external adversaries. Internally, compromised clients participating in the FL process may attempt to poison the global model by injecting manipulated updates during local training. Such updates could introduce backdoors or degrade model accuracy, as demonstrated in \cite{bagdasaryan2020backdoor}. The immutability of blockchain technologies, such as Ganache, amplifies this risk by permanently recording malicious updates, enabling persistent attacks that are difficult to reverse \cite{desai2021blockfla}.
The use of IPFS for storing models adds another layer of vulnerability. If a poisoned model is uploaded and propagated, its content-addressable hash becomes permanently accessible within the distributed network.
Beyond poisoning attacks, adversaries may exploit the underlying smart contract infrastructure through blockchain-specific threats such as re-entrancy and front-running attacks. 

The reputation-based voting mechanism introduces risks of adversarial coordination. Attackers might create Sybil identities to gain disproportionate voting power, artificially approving substandard models by forcing them to meet threshold scores.  If the reputation-based smart contract fails to detect attacks involving colluding clients artificially inflating their reputation scores to approve malicious models, the flawed model could be widely adopted \cite{luu2016making}. Sybil attacks in decentralized governance systems, where adversaries create fake identities to manipulate voting outcomes, have been studied extensively in blockchain and FL contexts. \cite{kang2019incentive} proposed a reputation-based blockchain framework for FL that penalizes malicious actors but acknowledges vulnerabilities to Sybil collusion if identity verification is weak. Similarly, \cite{fung2018mitigating} surveyed Sybil attack vectors in FL, highlighting how decentralized voting mechanisms (like those in smart contracts) are particularly susceptible to fake identities overwhelming consensus processes. 


Even with internal model hosting, external adversaries can still target it indirectly. Adversaries might reverse-engineer the model by analyzing input-output pairs stored on the blockchain, replicating techniques from \cite{tramer2016stealing}, who extracted models via prediction APIs. Adversarial evasion attacks involving crafting subtle perturbations to input data (e.g., temperature readings), could bypass reputation checks and force incorrect forecasts, as theorized in \cite{makhzani2015adversarial}.

\subsection{Security Assumptions}
\begin{itemize}
\item The decentralized architecture inherently mitigates large-scale DDoS attacks by distributing the workload across multiple nodes. Additionally, since the IPFS node is deployed locally, the threat of external attackers launching DDoS attacks on IPFS gateways is considered negligible in this deployment scenario. 
\item The project is deployed and operated within a secure local environment, with no external network exposure. As a result, the risk of man-in-the-middle attacks is considered negligible. Therefore, the use of encrypted communication protocols, such as Transport Layer Security (TLS), for data transmission is not required in this specific deployment scenario.
\end{itemize}

\section{System Design and Architecture}
\subsection{High-Level System Overview}
\begin{figure*}[htbp]
    \centering
    \includegraphics[width=1.04\textwidth]{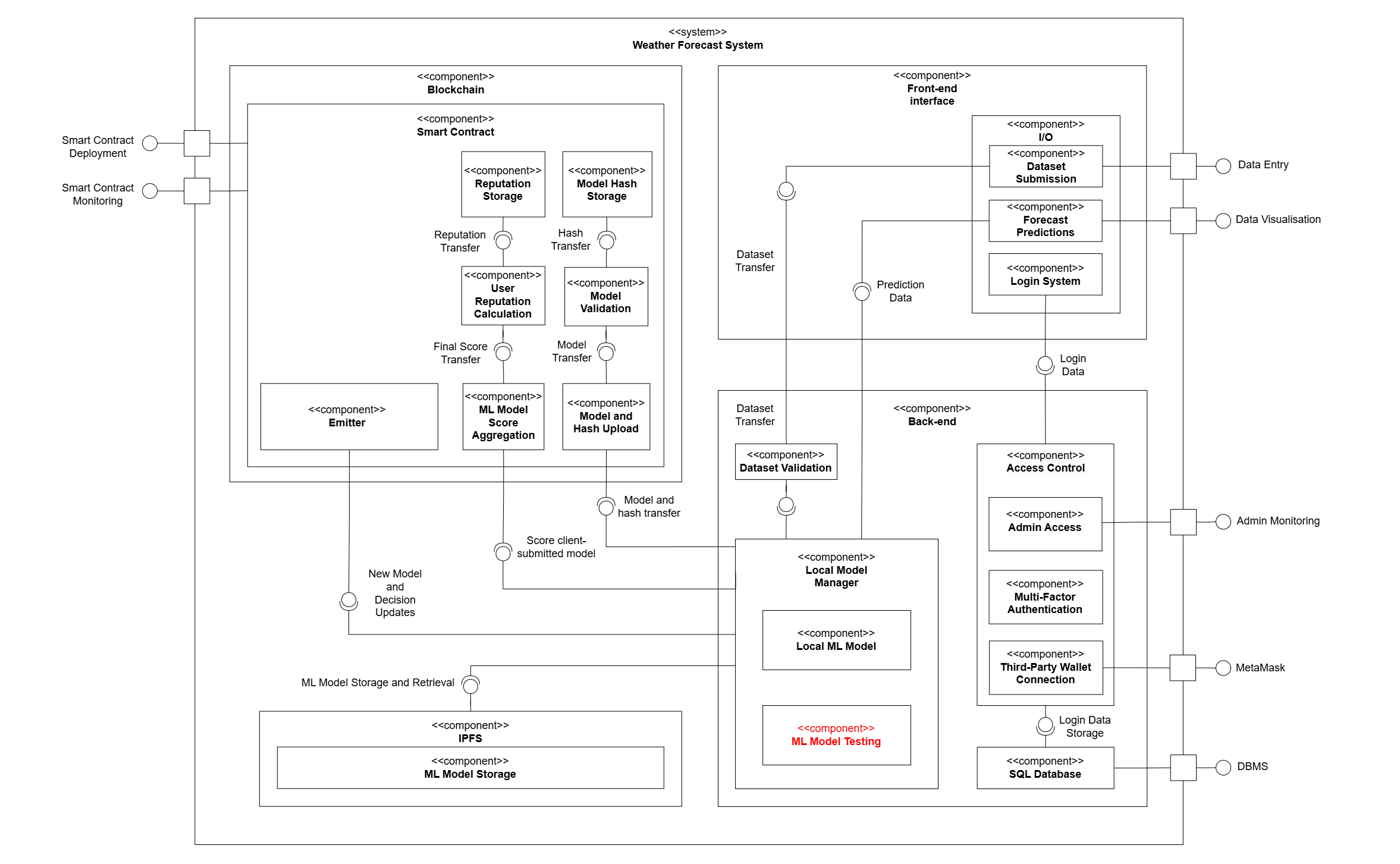}
    \caption{Overall system component diagram that contains the four key components, and the various connections and interconnections between these components. External interfaces to users, clients, and administrators are shown on the left side of the diagram (smart contract interfaces) and the right side of the diagram (python-flask interfaces). ML model testing (highlighted in red) is not implemented yet. Within the diagram, circles represent the interfaces provided by a connected component, and sockets represent the interfaces required by a connected component.}
    \label{fig:SystemComponentDiagram}
\end{figure*}

The implemented system comprises four core components: a smart contract, a Python-Flask backend, a web-based frontend interface, and an IPFS based storage layer. As shown in Figure~\ref{fig:SystemComponentDiagram}, these modules interact to support secure and decentralized weather forecasting via FL. The smart contract manages the reputation-based model scoring and voting. The backend, implemented in Flask, handles model training, inference, and client interactions. The frontend provides a graphical interface for users, clients, and administrators. Trained models are stored on IPFS, while their associated CIDs are stored on-chain to minimize gas costs and exposure.

\subsection{Python-Flask System Overview}

\begin{figure}[htbp]
    \centering
    \includegraphics[width=0.498\textwidth]{ 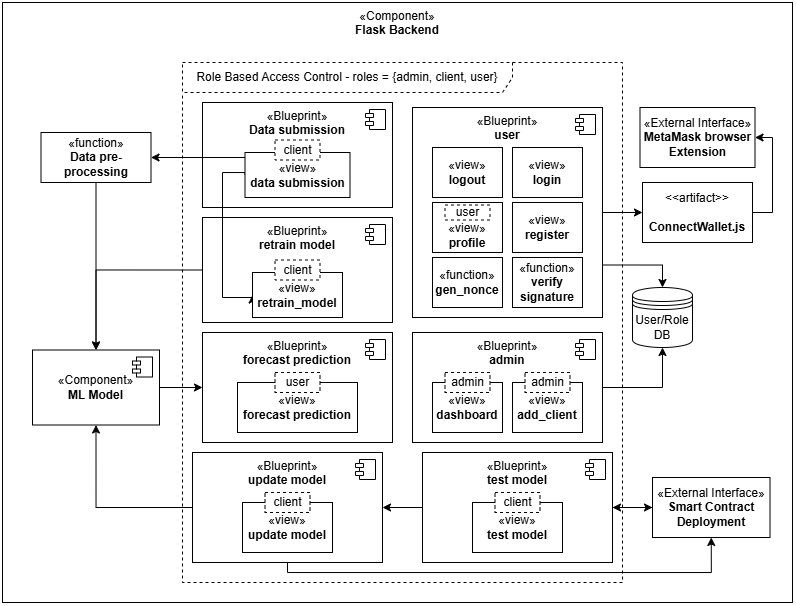}
    \caption{Component diagram that represents the overall structure of the Flask backend application, which enables a user to view and interact with the ML model through the smart contract via the Flask website.}
    \label{fig:backendComponent}
\end{figure}

The backend component diagram, as shown in Figure \ref{fig:backendComponent}, represents the structure of the Flask application, identifying the specific components and functions that are covered by role-based access control and the connections between these components. Each blueprint within the diagram is labeled with the role that has access to that specific page. Arrows show the connection between components, including the interaction with the external components such as the MetaMask browser plugin and the smart contract deployment.

\subsection{Stakeholders and Interactions}
\label{sec: stakeholders}


The system incorporates several roles with varying levels of authority. At the base level are Users, who are limited to utilizing local machine learning models to generate weather forecast predictions. Clients possess additional privileges, including participation in model retraining, submission, and engagement in the consensus-based model scoring mechanism. Admins hold elevated permissions that enable them to register Users as Clients. These Admins are initially assigned the highest reputation scores, as they are presumed to be trusted individuals selected by the system's developers. The Owner is responsible for deploying the smart contract and has the overarching authority to designate system administrators.

The backend API manages user accounts and their associated roles using an SQL database. Role-Based Access Control (RBAC) is enforced within the Flask application by querying the database to determine each user's role and making access control decisions accordingly. 
Externally, some developers may wish to build their front-end interface to interact with our smart contract. Data providers take the form of users who retrain the ML model, and the initial dataset providers~\cite{weather_analysis_kaggle,weather_dataset_kaggle}. 

\subsection{Federated Learning Pipeline, Model Updates and Model Score Aggregation}

\paragraph{Local Model Training on Distributed Weather Data Sources}
FL follows a decentralized approach where local models are trained on edge devices or distributed weather data sources without sharing raw data \cite{mcmahan2017communication}. Local model training in this FL setup involves using sktime, an ML framework specialized for time series forecasting \cite{loning2019sktime}, and Long Short-Term Memory (LSTM) networks, a type of Recurrent Neural Network (RNN) well-suited for sequential data \cite{brownlee2017long}. The goal is to train a weather forecasting model locally before storing and sharing it via IPFS. 
The local model training process is performed using Sktime's LSTM-based neural forecasting model (NeuralForecastLSTM) and an LSTM-based classification model (LSTMFCNClassifier). The weather dataset consists of features such as temperature, humidity, wind speed, visibility, and atmospheric pressure, which are essential for accurate weather predictions \cite{makridakis2018statistical}. 
By making use of these ML approaches, the system achieves innovation by making use of advanced ML techniques that go beyond traditional ML approaches like random forest or gradient descent.

The dataset undergoes pre-processing steps, including handling missing values, label encoding for categorical variables, and dataset normalization using Z-score standardization to ensure stable training. The dataset is then split into training and testing sets, ensuring that future data points are not leaked into the training phase \cite{hyndman2018forecasting}.

The regression model is trained using a regression LSTM forecaster, which is well-suited for time-series forecasting due to its ability to capture temporal dependencies \cite{hochreiter1997long}. The model is trained using historical weather data, predicting temperature for the forecast horizon, and evaluated using Mean Absolute Error (MAE), Root Mean Squared Error (RMSE), and Mean Absolute Percentage Error (MAPE). 

For classification tasks, an LSTM classifier is used to predict categorical weather attributes. One-hot encoding is applied to categorical variables, ensuring compatibility with the neural network model \cite{goodfellow2016regularization}. The model is evaluated using accuracy and F1-score. 
The trained models are saved using Joblib for future inference. Joblib, a specialized Python framework, is useful for serializing and deserializing models, which allows local clients to store trained models and reuse them without retraining \cite{joblib}. This approach supports privacy-preserving weather forecasting by ensuring that data remains on local devices while still benefiting from advanced ML techniques~\cite{li2020federated}.

\paragraph{Model Storage on Inter-planetary File System}
Upon local training, the model is uploaded to IPFS via its HTTP API. IPFS assigns a unique Content Identifier (CID), computed as a cryptographic hash of the model file, ensuring tamper resistance \cite{IPFS_Docs}. This CID is submitted to the smart contract, which references it for model validation and scoring. Any model alteration results in a different CID, enabling immutable and verifiable storage. Clients performing retraining operations also submit updated models through this process \cite{batchu2021,ipfscontentaddressed}. This process ensures that all model updates are securely stored and tracked on IPFS, maintaining integrity and providing a reliable source of truth for model performance and consensus-based updates \cite{ipfscontentaddressed}.


\paragraph{Consensus-based Model Score Aggregation Using Smart Contracts}
Clients with at least 10 reputation points are eligible to evaluate submitted models. Once sufficient scores are collected, i.e., when the total voting reputation exceeds a threshold, the smart contract computes a final score as a reputation-weighted average. Based on this score, the model submitter's reputation is updated. If the model outperforms the current primary model, it is promoted to primary status and disseminated to clients for inference.

\subsubsection{Prevention of adversarial attacks}
\label{sec: Attacks}
\begin{itemize}
    \item Smart contract functions are implemented following the checks-effects-interactions design pattern. This sequence ensures that all validation checks occur before any state modifications, and that external calls are deferred until the end of the function. As a result, even if an adversary attempts to re-enter the function mid-execution, altered contract states will invalidate subsequent invocations, rendering the attack ineffective.
    \item Sybil resistance: All clients begin with zero reputation and must accumulate a minimum of 10 reputation points before participating in model scoring. This significantly raises the cost and complexity of orchestrating a Sybil attack, as adversaries would need to operate numerous independent clients over time while avoiding detection.
    %
    
    \item Data poisoning protection: Achieved by a consensus verification of client-submitted ML models. If a client submits a model with poor accuracy, other clients within the consensus-based model scoring system will assign a low score to it, and it will therefore not reach primary status and will never be used for forecast predictions.
    \item Front-running attack prevention: The influence of any single client is limited by their reputation score. Even if an attacker attempts to front-run a transaction by prioritizing their vote via increased gas fees, their voting impact remains proportional to their reputation.
\end{itemize}

\subsection{Access Control}
\paragraph{Flask Backend}
The backend enforces RBAC to regulate access based on user roles: admins (manage logs and register clients), clients (submit and score models), and users (access forecasts). Access checks are implemented using middleware, and unauthorized requests return a 403 error and are logged. Key events such as registrations, sign-ins, submissions, and rate limit violations are recorded and viewable via a secure admin dashboard.

\paragraph{Smart Contract}
Smart contract functions use Solidity's require statements to enforce access control, verifying caller identity via wallet addresses. This ensures that only authorized roles can invoke critical operations. Impersonation is infeasible without access to a user’s private key. Future improvements include syncing admin roles between the contract and Flask backend to maintain consistent authorization across components.

\section{Implementation}
The implemented system integrates multiple technologies to enable FL, blockchain-based security, and decentralized model verification. The key components include an FL Framework, a Blockchain Platform, and API Integration. SKtime is used for time-series forecasting, while LSTM networks are employed for predictive modeling. An Ethereum smart contract ensures decentralized governance and model verification. The Flask framework serves as the backend, enabling seamless communication between FL models and blockchain smart contracts.

\subsection{Federated Learning Model Training Process}
\paragraph{Data Partitioning and Distributed Training Setup}

Weather datasets are spread across multiple client devices, eliminating the need for centralized data collection. Each client trains a local LSTM model using their private dataset, ensuring that raw data remains secure and never needs to be shared. Instead of directly exchanging data, clients submit their retrained updated models to IPFS, which are then voted in as the primary model used by all users to improve overall model performance.
\paragraph{Validation and Non-Malicious Contributions}

A differential privacy mechanism ensures that no single client can access sensitive data from others, maintaining data security. Anomaly detection techniques are used to identify and filter out malicious model submissions that attempt to introduce adversarial noise or backdoors. Clients who submit unreliable models receive negative reputation scores, helping to prevent dishonest behaviour and maintain trust within the system.


\subsection{Flask Backend Implementation and API Integration}
The Flask \cite{flask} web application framework runs a backend server for the web interface of the decentralized forecasting system. It facilitates interaction with ML models stored in IPFS and records their CIDs in the smart contract deployed on Ganache. This setup enables clients to run an instance of the framework and interact with smart contract functions, such as the consensus-based model scoring mechanism for model updates.
The Flask backend also serves as a web server, integrating with the front-end interface where users can submit training data and receive predictions. The front-end uses Jinja2 templating to pass data from the backend to HTML pages, while JavaScript handles user interactions, including connecting a user’s MetaMask wallet via a browser extension. For styling, the Bootstrap CSS framework \cite{bootstrap} is used and is integrated through a Cloudflare Content Delivery Network (CDN) link.

\subsection{Smart Contract Implementation}
The smart contract for the project (smart\_contract.sol) was developed using the Solidity programming language, using the Remix Desktop IDE \cite{remixpro13:online}. Using Git version control allowed for collaborative development on a single smart contract between multiple team members, although should the project expand into further features, a segmentation of features into different smart contracts that interact with one another is an area identified for future development to reduce gas fees. The smart contract is split into three different types of entities: variables, events, and functions. Events allow for the smart contract to report new updates to clients listening to the corresponding blockchain transactions, and functions act as an interface to allow for clients to interact with the smart contract (e.g., to submit a new model CID or submit a score for a client-submitted model).



\subsection{Smart Contract Deployment}

The smart contract (smart\_contract.sol) was developed using the Solidity programming language within the Remix Desktop IDE \cite{remixpro13:online}. It includes variables, events, and functions that collectively implement role-based access control, model scoring, and reputation management. Events are emitted on critical state changes, such as model submissions or voting outcomes, to enable real-time client updates. Ganache was chosen for deployment as it offers a local blockchain environment with built-in wallet generation and a persistent workspace. While public Ethereum testnets offer a more authentic experience, their reliance on token faucets and frequent resets presents challenges for iterative testing. Ganache provided a stable, fast, and self-contained environment for development and testing without incurring deployment costs.

\subsection{Decentralized Storage and Integration}
All trained models are stored off-chain using the IPFS content-addressable storage system. IPFS ensures that each model can be verified and retrieved using its cryptographic CID. Any modification to a model results in a different CID, enabling tamper detection and maintaining integrity \cite{IPFS_Docs, batchu2021, ipfscontentaddressed}. This design decouples heavy model files from on-chain storage, reducing gas fees and improving scalability.

\section{Evaluation}

%

The implemented system meets its stated goal of providing a decentralized, privacy-preserving weather forecasting framework that integrates FL, blockchain-based validation, and decentralized storage via IPFS. Clients successfully retrain models on local data and contribute to the consensus-driven model selection process using smart contracts. The project successfully combines FL and blockchain technology as specified in  Section~\ref{sec: intro}.

\subsection{Robustness Against Adversarial Attacks}
Each attack prevention highlighted in section \ref{sec: Attacks} contributes to the overall security and integrity of the system:
\begin{itemize}
    \item Re-entrancy attack prevention: Ensures smart contract function execution correctness, especially concerning smart contract state changes.
    \item Sybil attack resistance: Ensures the reliability of consensus-based voting, removing the possibility of final model score skewing by an adversary.
    \item Data poisoning protection: Ensures that only models that have improved performance over the current model will be accepted as replacements.
    \item Front-running attack prevention: Preserves the order of transactions submitted by clients, which is especially relevant for the implemented model scoring system since future model scores are ignored once a reputation threshold from score-submitters is achieved.

\end{itemize}

\subsection{Performance Benchmarking}
The main performance benchmarks we consider for ML are the time to run and the scores across 5 different scoring metrics. Three of these metrics measure the prediction quality of the regression model: Mean Absolute Error (MAE), Root Mean Square Error (RMSE), and Mean Absolute Percentage Error (MAPE). The other two metrics measure the classification model (Accuracy and F1 score). Incorporating several metrics here allows us to see how the models perform against slightly different scoring criteria.  
The time performance considered three files: File verification (ensuring files are safe and correctly formatted to parse), Data pre-processing (Formatting the raw data into an ML format), and the model. The timing results for these files are shown in Table~\ref{tab:timing_results}.


\begin{table}[h]
\centering
\scalebox{1.19}{
\begin{tabular}{|l|c|p{8cm}|}
\hline
\textbf{Process} & \textbf{Time (s)} \\ \hline
File verification & 0.1  \\ \hline
Data preprocessing & 0.05  \\ \hline
Model & 1213  \\ \hline
\end{tabular}
}
\caption{Runtime of file verification, data pre-processing, and model execution.}
\label{tab:timing_results}
\end{table}

It is worth noting that although the model took a long time to run, this was primarily due to the training process with a relatively large dataset. Once training was completed, the prediction was run sequentially for each variable within the dataset to test the model's predictive capabilities. This was also very time-consuming, as an instance of the model (either classification or regression) is run each time on the different targets. For quick testing and a balance of runtime vs. accuracy, this was throttled to 50 epochs of training.  

The scoring performance returned the average results for the scoring across the different metrics. The results obtained are shown in Table~\ref{tab:performance_metrics}. Note that these values may vary slightly if the model is run at a different time.

\begin{table}[h]
\centering
\scalebox{1.19}{%
\begin{tabular}{|l|c|}
\hline
\textbf{Metric} & \textbf{Value} \\ \hline
MAE & 1.9543 \\ \hline
RMSE & 2.0480 \\ \hline
MAPE & 1.5007 \\ \hline
Accuracy & 0.5 \\ \hline
F1 & 0.5 \\ \hline
\end{tabular}%
}
\caption{Performance metrics of the model.}
\label{tab:performance_metrics}
\end{table}

These results do not clearly illustrate the exact performance for each target variable, but instead provide a higher-level overview of the overall model performance. In some cases, the model predicted almost perfectly, however, there were only 2 or 3 possible options, and heavily weighted in favour of another. In other cases, the model struggled to predict accurately, where there were several possible outcomes that were minutely different (e.g., Weather Summary).

We observed that in the split functionality, where the data was parsed into training and testing sets, there is a hard-coded train call specifically on the Temperature column. This was originally built for testing purposes, but upon fixing this, any efforts to replace this hard-coded temperature with the target variable to predict vastly skewed the results. Scores for MAE and RMSE fluctuated unreliably, sometimes in the thousands. Due to this, we retained this direct ``Temperature'' training, despite acknowledging that it remained from testing code, in the interest of providing more accurate predictions.

\subsection{Smart Contract Evaluation}
\label{sec: smart_eval}
Evaluation of the smart contract was achieved through unit testing. This was performed through Truffle (which was also used for deployment). Testing was performed on each function, with tests covering normal usage as well as cases of smart contract misuse. These tests were written in a JavaScript file, which was run regularly during the development of the contract to check smart contract development progress. 20 unit tests were used to evaluate the correctness of the smart contract. After some fixes were implemented, all 20 tests passed. Issues that were fixed during testing ranged from clients not being added correctly to issues of reputation scores not being calculated for scores below certain thresholds.

Smart contract testing focused on correct function execution, access control, and prevention of unauthorized actions. During deployment, an initial failure occurred due to Solidity version incompatibility (0.8.26), which was resolved by downgrading to 0.8.13. Unit tests verified that only authorized wallets could execute critical functions. The setMain() function successfully restricted updates to the original deploying wallet. Similarly, the addPrivileged() function enforced strict user additions, blocking unauthorized inputs, and preventing duplicate entries.
Functionality validation for addClient() initially failed due to a bug where the msg.sender was incorrectly assigned as the new user instead of the intended address. After correction, further tests confirmed the function's reliability. The storeHash() function was also tested to ensure that only registered clients could submit data, effectively blocking unauthorized or duplicate submissions

The castMLScore() function, responsible for model scoring and client reputation updates, encountered an integer overflow when processing votes, leading to inaccurate score calculations. This issue was resolved through adjustments in the calculate reputation function. Additional tests verified that low-reputation clients were unable to influence final scores, safeguarding against manipulation attempts.

Threat simulations for the smart contract included re-entrancy attacks, which were successfully mitigated by implementing state-lock mechanisms. Tests for Sybil attacks showed that multiple low-reputation wallets could not manipulate the system due to enforced voting weight restrictions. Furthermore, stress tests ensured that gas limits were optimized, preventing DoS vulnerabilities.

\begin{figure}[htbp]
    \centering
    \includegraphics[width=0.94\linewidth]{ 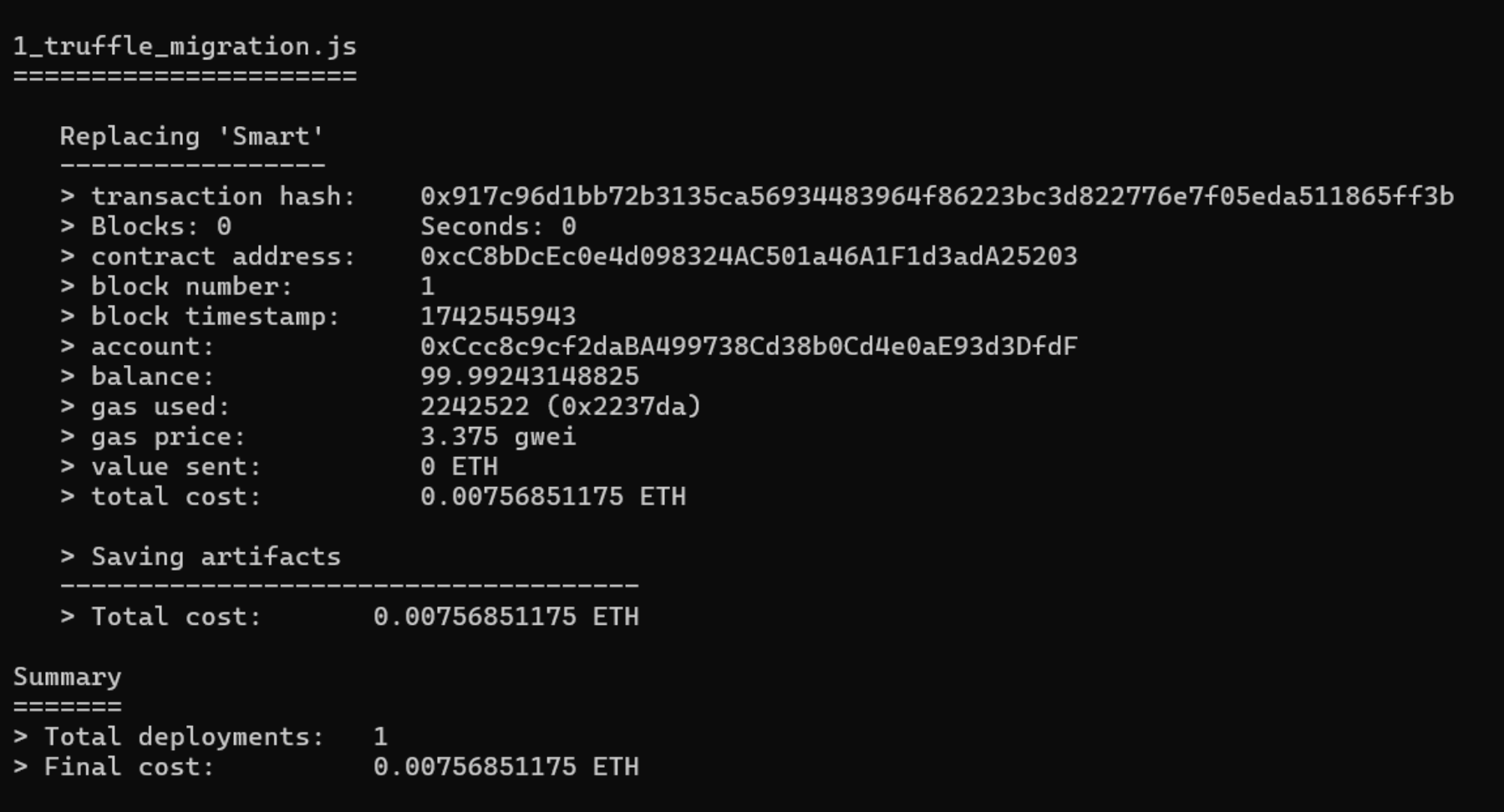}
    \caption{Deployment Gas usage before gas optimization (commit aa36409)}
     \label{fig:beforeoptimization}
     \vspace{-3mm}
\end{figure}

\begin{figure}[htbp]
    \centering
    \includegraphics[width=0.94\linewidth]{ 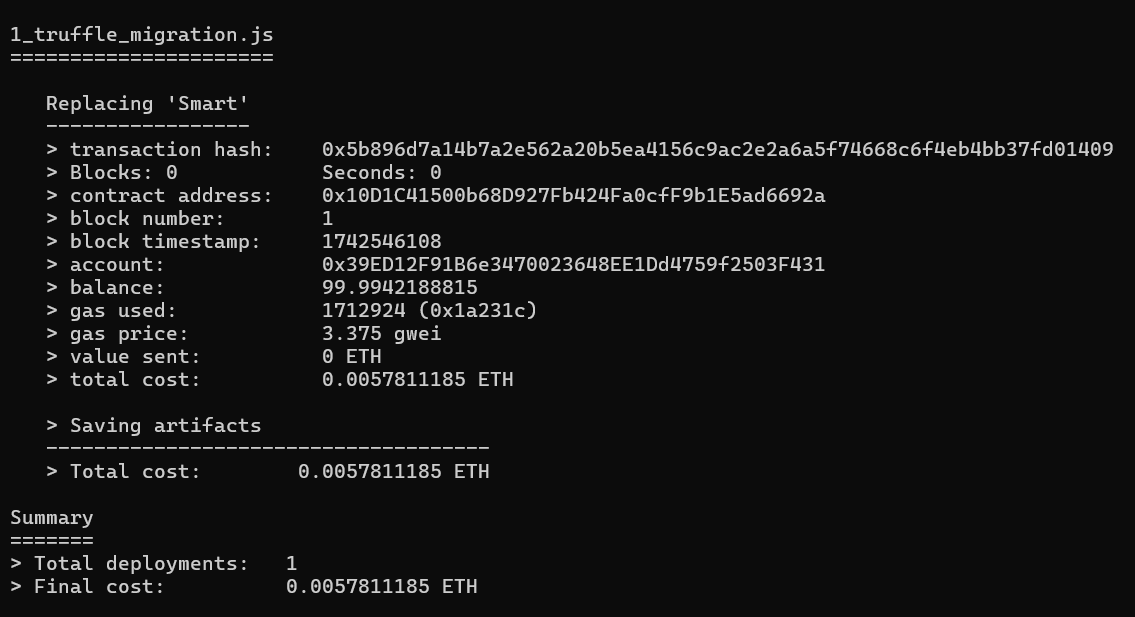}
    \caption{Deployment Gas usage after gas optimization (commit a237735)}
     \label{fig:afteroptimization}
\end{figure}

Initial steps were taken towards the end of development to optimize the gas usage of the project’s smart contract. Firstly, any unused variables, functions, and parameters were removed. An array was swapped to a mapping, as suggested by \cite{young2025} \cite{Solidity2:online}. Repeated 'require' clauses have been aggregated into function modifiers that reduce redundant code. Following gas optimization and bug fixes, Figure \ref{fig:beforeoptimization} and Figure \ref{fig:afteroptimization} demonstrate that the deployment gas usage was reduced from 2242522 gas units (7,568,511.75 gwei) on commit aa36409, to 1712924 gas units (5,781,118.5 gwei) on commit a237735, a reduction of 23.6\% (rounded up to one decimal place), with a gas price of 3.375 gwei. 

\subsection{Flask Backend Evaluation}

Security analysis for the Flask backend covered access control, injection vulnerabilities, security misconfiguration, and authentication failures. Tests for broken access control demonstrated that unauthorized users could not bypass security measures through direct object reference manipulation or API testing with Hoppscotch. Injection testing included SQL, OS command, and NoSQL injections via SQLmap and Burp Suite, all of which were successfully blocked, confirming the contract’s resistance to common attack vectors.
To detect security misconfigurations, Nikto and Nmap scans were performed, verifying that no outdated software, exposed services, or missing security headers posed risks. Authentication mechanisms were tested against brute-force attacks using Hydra, confirming that rate limiting and account lockout policies effectively prevented unauthorized access.

\subsection{The Result Overview}

\begin{figure}[htbp]
    \centering
     \includegraphics[width=\linewidth]{ 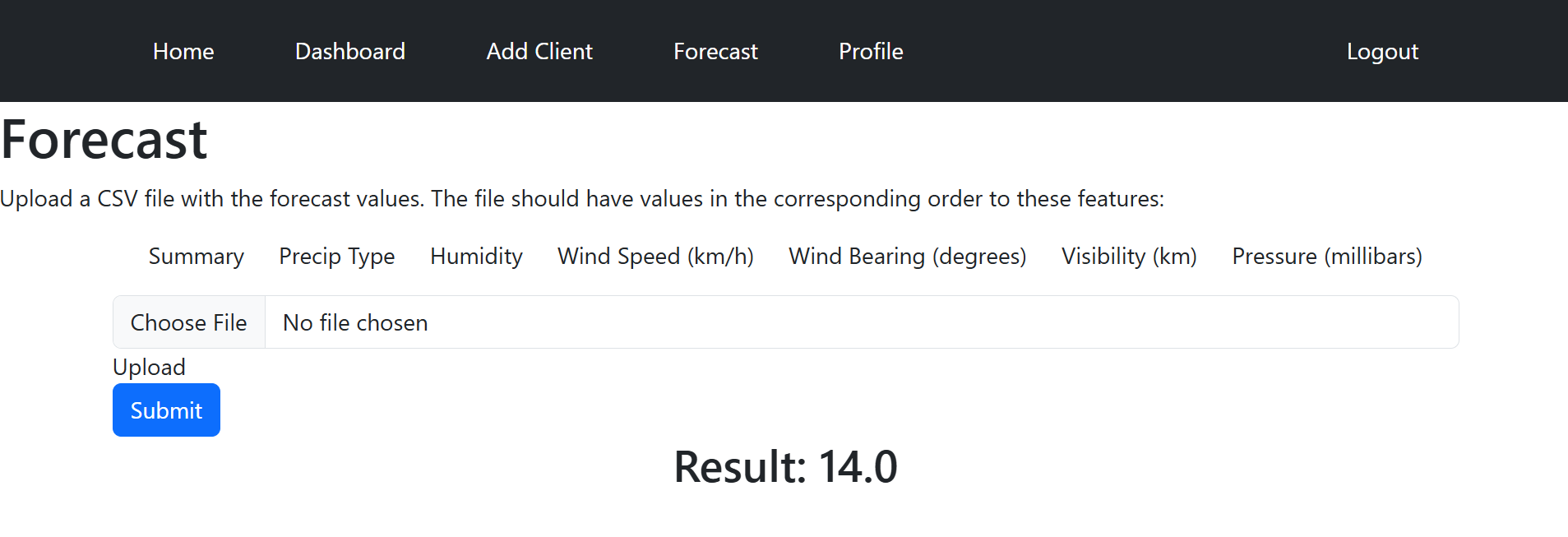}
     \caption{Display of the forecast page where a user can submit a dataset that includes values to pass to the regression ML model. A prediction on the temperature for the user's dataset will be generated and rendered in the result.}
    \label{fig:forecast-page}
 \end{figure}

To evaluate the system's functionality, a user submitted a weather dataset to a locally pre-trained machine learning model. As shown in Figure \ref{fig:forecast-page}, the model produced a forecasted temperature of 14°C, which was subsequently submitted to the blockchain-based validation layer. The model was independently developed and not shared across clients, thereby maintaining the decentralized integrity of the framework.

\section{Conclusion}

In this paper, we presented a decentralized weather forecasting framework that integrates federated learning with blockchain technology to enhance privacy, trust, and system resilience. By enabling local training on client devices and using blockchain-based smart contracts for model validation, our system helps mitigate risks related to data tampering, model poisoning, and centralized failure. The use of IPFS for off-chain storage further ensures scalability and tamper resistance while minimizing on-chain storage cost. Experimental results show the feasibility and robustness of our approach in real-world scenarios.

However, various challenges remain. Future work could focus on improving predictive performance through advanced hyperparameter tuning and expanding the system's applicability to a broader set of forecasting tasks. One of the limitations of the proposed system is its reliance on a centralized owner to assign administrative roles and register new clients, which introduces potential trust vulnerabilities. Future work can address this by adopting a consensus-based mechanism for client onboarding and admin governance.

\section*{Acknowledgments}
 This project received funding from UK National Edge AI Hub under grant number EP/Y028813/1 and EP/W003325/1. 
\bibliographystyle{plain}
\bibliography{ref}


\end{document}